# Towards spoken dialect identification of Irish

*Liam Lonergan[1], Mengjie Qian[2], Neasa Ní Chiaráin[1], Christer Gobl[1], Ailbhe Ní Chasaide[1]*

[1]Phonetics and Speech Laboratory, School of Linguistics, Speech and Communication Sciences, Trinity College Dublin, Ireland
[2]Engineering Department, Cambridge University, UK

mq227@cam.ac.uk, {llonerga, nichiarn, cegobl, anichsid}@tcd.ie

## Abstract

The Irish language is rich in its diversity of dialects and accents. This compounds the difficulty of creating a speech recognition system for the low-resource language, as such a system must contend with a high degree of variability with limited corpora. A recent study investigating dialect bias in Irish ASR found that balanced training corpora gave rise to unequal dialect performance, with performance for the Ulster dialect being consistently worse than for the Connacht or Munster dialects. Motivated by this, the present experiments investigate spoken dialect identification of Irish, with a view to incorporating such a system into the speech recognition pipeline. Two acoustic classification models are tested, XLS-R and ECAPA-TDNN, in conjunction with a text-based classifier using a pretrained Irish-language BERT model. The ECAPA-TDNN, particularly a model pretrained for language identification on the VoxLingua107 dataset, performed best overall, with an accuracy of 73%. This was further improved to 76% by fusing the model's outputs with the text-based model. The Ulster dialect was most accurately identified, with an accuracy of 94%, however the model struggled to disambiguate between the Connacht and Munster dialects, suggesting a more nuanced approach may be necessary to robustly distinguish between the dialects of Irish.

## 1. Introduction

Automatic speech recognition (ASR) systems are generally built for the spoken 'standard', and their performance declines when the user speaks a dialect or a non-standard variety [1, 2]. This is a problem for a language like Irish, where there is no single spoken standard, but rather three major native dialects, namely Ulster (Ul), Connacht (Co) and Munster (Mu). , with many sub-dialects, as well the accents of learners and new speakers [3]. Therefore, it is a priority to cater for this variation from the outset of system building. A recent study finds unequal performance across dialects in Irish ASR with dialect-balanced training corpora [4], indicating that additional system building strategies may be necessary to reduce this disparity, such as the inclusion of modelling dialect as part of the ASR process.

The current paper outlines recent efforts in developing a spoken dialect identification (DID) system for Irish. Generally, language identification (LID) and DID tools act to route utterances to language-specific or dialect-specific ASR models, however due to the small scale of available Irish speech corpora, a pooling together of all data is necessary for training ASR systems. Nonetheless, exploiting the mutual information shared between the DID and ASR tasks is of interest and will be explored in future experimentation to reduce the disparity in Irish ASR performance across dialects. While this use of DID is not explored here, the current experiments constitute an important first step in developing DID for Irish.

This paper reports results from experiments in building an Irish spoken DID system. Two speech classification models are tested: XLS-R [5] and ECAPA-TDNN [6]. To determine whether text-based DID can be helpful in improving accuracy, an ECAPA-TDNN, the best performing speech classification model, is fused with a text-based DID model.

## 2. Background

### 2.1. Irish language

Irish is a highly inflectional, Celtic language and is recognised as the first language of the Republic of Ireland. Irish is endangered [7], despite being a compulsory subject in primary- and secondary-level education, and is spoken as a community language in small Gaeltacht regions scattered along the western seaboard. Colonization and increased prevalence of English contributed to the gradual decline of the language.

*2.1.1. Dialects of Irish*

As language loss spread westward, Irish-speaking communities became isolated from one another, resulting in little communication between Gaeltacht areas. However, the emergence of Irish language radio and television in recent times has facilitated greater exposure and mutual understanding among different dialects [8]. These dialects vary in many ways. Linguists have held different perspectives on the categorization of these dialects, with the consensus of three broad dialects as set out in the introduction, as well as, for example, O'Rahilly [9] suggesting only two groups: a northern grouping (Ul) and a southern grouping (Co and Mu).

*2.1.2. An Caighdeán Oifigiúil: standard written form*

An Caighdeán Oifigiúil, a standardised written form of Irish, was introduced in 1958 for internal use of the Translation Section of the Irish Government. However, its adoption by the Department of Education quickly established it as the dominant form for all official and semi-official purposes. It is important to note that the standard is artificial and does not reflect aspoken variety of the language. Irish-language text corpora are predominantly composed by texts written in the standard form, and dialectal texts often lack appropriate metadata tags.

## 2.2. Dialect bias in Irish ASR

A recent study [4] investigated how the proportional representation of dialects in the training set of an Irish ASR system can result in performance bias and showed that balanced corpora do not lead to equitable performance, with Ul consistently performing worse than the other dialects. An interesting, close relationship was revealed between Co and Mu dialects. Motivated by these results, the present experiments seek to develop a spoken dialect identification (DID) system for Irish, with a view to incorporating DID into the ASR pipeline to improve dialect performance equity. Interestingly, a similar trend is reveled in the present experiments, where Ul is markedly distant from the other two dialects, and Co and Mu are difficult to disambiguate.

## 2.3. Dialect identification (DID)

Language identification (LID) is the process of automatically identifying the language of speech or text. Dialect identification (DID) is a special case of LID that focuses on disambiguating between dialects within a language. Although many techniques can be applied to both LID and DID tasks, DID is often more challenging due to the greater linguistic divergence between languages compared to dialects within a language.

As stated in Section 1, the conventional application of DID, such as routing utterances to dialect-specific ASR models, is not feasible for Irish due to the small scale of available corpora. Nonetheless, DID can still be used to improve dialectal speech recognition by either using embeddings from DID systems as auxiliary acoustic features for an ASR model or by jointly training ASR and DID in a multi-task learning set-up. Beyond ASR, DID holds potential for future applications by providing dialect meta-data for unlabelled speech data. To improve accuracy in this scenario, the fusion of text-based dialect classification with ASR outputs is explored below.

## 3. Related work

The i-vector approach was previously the state-of-the-art method for a range of speech classification tasks, including language identification [10, 11]. Embeddings are extracted in a fixed-length fashion, and linear discriminant analysis along with cosine distance or logistic regression is used as the backend system for classification. Neural networks were subsequently explored for generating such embeddings for speech classification tasks, most notably the x-vector approach. Frame-level features are aggregated into utterance-level embeddings and are made robust through using extensive data augmentation techniques, such as adding noise and reverberation to the training data [12, 13]. Different loss functions were explored for training x-vector extractors and the angular softmax loss has been widely adopted [14]. Inspired by the x-vector extraction system, the developers of ECAPA-TDNN [6] made a series of enhancements to that architecture including Squeeze-Excitation Res2Blocks and multi-layer feature aggregation, resulting in improved performance. The ECAPA-TDNN architecture has seen state-of-the-art performance for many speech classification tasks [6]. Wav2Vec 2.0 has been shown to act as a powerful frontend for speech classification tasks [15]. Most recently, Conformer encoders have been explored for LID. The Oriental Language Recognition challenge series has focused on identifying closely related Asian languages in recent years. The winners of the 2021 challenge pretrained their LID system as a U2++ Conformer encoder-decoder ASR. The encoder was then finetuned for LID by adding an attentive statistics pooling layer followed by a linear layer to the output nodes. The pretraining stage allowed the encoder to learn phonetic information that was helpful in disambiguating between the language groups. This system was 66% more accurate than the team that placed second [16]. Most recently, Meta released a model from the Massively Multilingual Speech project, which can perform LID by stacking a linear classifier on top of the encoder for over 4,000 languages [17], using a model structure similar to the XLS-R model trained in the present experiments.

Due to the closer relationship between dialects of a single language than between different languages, DID approaches that leverage linguistic knowledge about the dialects can help to bolster performance. The Tibetan language Ao has three lexical tones and the tone assignment in lexical words acts as a dialect marker. [18] explores using excitation-source features which characterise the F0 contour for DID in Ao. Using these features in conjunction with MFCCs improves performance over using only MFCCs. [19] studies the potential of including prosodic features in Arabic DID. Using prosodic features alone, a classification accuracy of 72% was obtained and when combined with phonotactic features, accuracy is improved over a purely phonotactic-based classifier from 83.5% to 86.3%.

Text-based dialect classification has previously made use of character or word n-grams features [21, 22]. More recently, self-supervised models such as BERT [22] language models have been employed for DID of Arabic texts [23].

The low-resource constraint of Irish speech recognition requires efficient and effective use of available speech corpora. Although not examined in the present experiments, the joint modelling of accent/dialect identification and speech recognition has demonstrated improved performance in accented ASR [10,11,12]. This approach is appealing as it makes use of the mutual information shared between the two tasks to bolster performance efficiently. Additionally, the integration of accent/dialect embeddings as auxiliary acoustic features has been shown to improve accented ASR [13,14,15]. These will be explored in future experiments.

## 4. Experimental Set-up

### 4.1. Data

The objective of the experiments is to distinguish different dialects based on the characteristics present in the speech and text data; therefore, efforts were made to ensure that our datasets were dialect balanced.

Table 1: *Dialect-balanced corpora in these experiments. Size is duration of acoustic materials and number of words for text.*

| Data Type | Dataset | Size |
|---|---|---|
| Acoustic | Carefully read speech | 48h |
| | Spontaneous speech | 90h |
| | Total | 136h |
| Text | Historical corpus | 4.4m |
| | Spon. speech transcripts | 4.2m |
| | Ulster prose material | 0.3m |
| | Total | 9.0m |

#### 4.1.1. Acoustic data

Details of the acoustic training data used in these experiments can be seen in Table 1. This data draws from in-house, carefully read recordings including both synthesis data and field recordings using dialect-appropriate prompts, for which dialect information of speakers was collected. A spontaneous speech corpus of broadcast material provided by Foras na Gaeilge's New English-Irish Dictionary project [30], which is tagged with dialect information, is also used. Due to the necessity of balancing the corpora, only a subset of our corpus as used in this study. The training set consists of ~16h of read speech and 30h of spontaneous speech data per dialect totaling 136h. The validation and test sets are 1.5h and 2.5h respectively of read speech of native Irish speakers. Special attention was paid to balancing the gender and dialect representation in both the validation and test sets, to prevent the acoustic classifiers from overfitting on gender-specific information rather than focusing on dialect information.

*4.1.2. Text data*

As explained in Section 2.1.2, access to text corpora with salient dialect features, that is also tagged with dialect information, is limited. The first portion of the training set is the data used to train an Irish text-based dialect classifier *canúint* [1], which consists of dialectal texts written between 1900-50 drawn from the Historical Irish Corpus of the Royal Irish Academy [31]. These texts were written before the introduction of the standard written form, so dialectal features are more marked than in modern texts. The second dataset is drawn from the transcriptions of the spontaneous speech corpus [30] used as acoustic data. This corpus is rich in natural, modern conversational data, and the transcriptions generally match the speech quite closely. As shown in Table 1, extra prose material of Ulster Irish was supplemented to ensure a balanced corpus. 90% of this data was chosen randomly for training and the remainder is used as a validation set in training.

### 4.2. Models

*4.2.1. Acoustic models*

Two acoustic models are tested in these experiments to classify dialects. The starting point of acoustic DID for Irish was the multilingual version of wav2vec 2.0 [32], XLS-R [5], which was pretrained using 436k hours from 128 different languages. The data used to pretrain this model contains < 10 hours of Irish data, a fraction of the overall dataset. In initial experiments, a variety of classification heads were tested, and the most successful approach was adopted from ARMBL[2], a community of researchers working on Arabic NLP. In this method, the frame-level embeddings are initially reduced from 1,024 to 128 with a bottleneck layer. The resulting reduced embeddings are concatenated together and projected using a hyperbolic tangent (tanh) activation function. The outputs are connected to the final layer, with a unit for each classification label (dialect). Additionally, freezing the XLS-R layers was tested and it was found that allowing all weights to be updated during training led to the best performance. This model was trained using the HuggingFace transformers framework [33].

The second model employed for acoustic-based dialect identification is ECAPA-TDNN [6], using 60-dimensional MFCCs as acoustic features. Two models are used in these experiments: a model trained from scratch and a pretrained ECAPA-TDNN model trained for a language identification task using VoxLingua107 dataset [34], a dataset comprised of 6.6k hours from 107 different languages. The pretrained model was finetuned for Irish DID by replacing the final output layer with 3 nodes. These models were trained using the SpeechBrain framework [35].

*4.2.2. Text dialect identification model*

The model used for text-based dialect identification is a classification head built on top of gaBERT [36], a BERT language model pretrained using 171M word tokens. The BERT model is made of stacks of transformer blocks to predict the identity of masked tokens and to predict whether two sequences are contiguous. A linear layer with dropout is added on-top of the gaBERT model and the model is finetuned for text-based dialect identification.

*4.2.3. Model fusion*

To combine the acoustic and text-based identification models, first the dialect probabilities of the text model for each validation set utterance transcription are extracted. A series of weights $\lambda$ were then tested according to this formula:

$$P_{dialect} = (1 - \lambda) * P_{acoustic} + \lambda * P_{text} \quad (1)$$

where $P_{acoustic}$ and $P_{text}$ correspond to the output probability scores for the acoustic and text-based models respectively. To find an optimal value for $\lambda$, a grid search was conducted with values between 0 and 1 with intervals of 0.1 on the validation set. The best performing fused system is used for the final evaluation on the unseen test set.

## 5. Results

### 5.1. Acoustic-based DID experiments

Table 2: *Classification accuracy of acoustic DID systems*

| Model | Accuracy |
|---|---|
| XLS-R | 55% |
| ECAPA-TDNN | 64% |
| Pretrained ECAPA-TDNN | 73% |

Initial acoustic-based DID experiments trained a simple classification head on XLS-R 300m, as detailed in Section 4.2.1. Despite the vast amount of speech this model has been exposed to in pretraining (436k hours), this model performed poorly for Irish DID, achieving 55% accuracy on the held-out test set. From the results it was clear that the system had difficulty disambiguating between the Co and Mu dialects and less trouble identifying Ul, which echoes the results from our dialect bias experiments outlined in Section 2.2

Due to the poor performance of the XLS-R system, we experimented with the ECAPA-TDNN architecture using two approaches: firstly, a model is trained from scratch, and secondly a pretrained ECAPA-TDNN model that was trained for an LID task with 107 languages is finetuned for Irish DID. The initial system had an accuracy of 64%, an improvement of 9% when compared with the XLS-R approach. From the confusion matrix in Figure 1(a), we can see again this model also had difficulty disambiguating between the Co and Mu dialects, whereas the Ul dialect was more easily recognized. Finetuning the pretrained ECAPA-TDNN boosts the accuracy by ~9% absolute compared with training the model from scratch. As can be seen in Figure 1(b), the same cross-dialect trend in performance between the dialects can be seen, however performance for Ul is dramatically improved.

---

[1] https://github.com/kscanne/canuint/

[2] https://github.com/ARBML/klaam

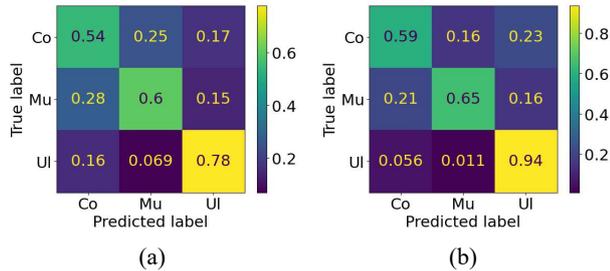

(a)          (b)

Figure 1: *Confusion matrix of (a) ECAPA-TDNN trained from scratch and (b) the finetuned ECAPA-TDNN model for Irish DID, both evaluated on the unseen test set.*

In Figure 2, embeddings from both XLS-R (left) and the best performing ECAPA-TDNN model (right) are visualised. These embeddings undergo two rounds of dimensionality reduction: the number of dimensions is initially reduced to 50 using principal component analysis (PCA), followed by t-distributed stochastic neighbour encoding (t-SNE) to further reduce the number of dimensions to 2. The XLS-R embeddings are extracted at the frame-level, and then averaged across an utterance to get an utterance-level representation before dimensionality reduction. In contrast, ECAPA-TDNN embeddings are extracted directly at an utterance-level and are not averaged. The ECAPA-TDNN embeddings exhibit clearer clustering of the dialects, highlighting the discriminative power of this approach.

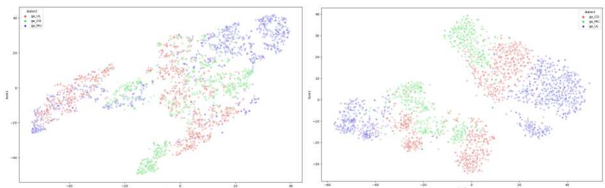

Figure 2: *Reduced model embeddings per utterance of unseen test set extracted from XLS-R (left) and ECAPA-TDNN (right) models. Co is marked in red, Mu is marked with green, and Ul is marked with blue.*

### 5.2. Text-based DID experiments

At the end of finetuning the gaBERT model for Irish text-based DID, the model had an accuracy of 87% with the text-corpus test set, however there is a significant disparity between the type of text the model is exposed to during training compared with the domain of text typically found in the transcriptions of our speech corpora. When tested on the transcriptions of the acoustic test set, the model had an accuracy of 54%.

### 5.3. Fusion of the two systems

The pretrained ECAPA-TDNN finetuned for Irish DID was fused with the text-based model. To find an optimal fusion weight, the output logits of the validation set for each model were combined according to the formula in Section 4.2.3.

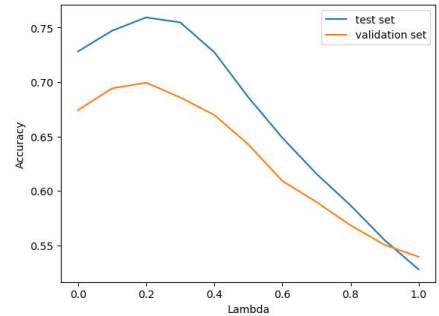

Figure 3: *Accuracy of fused model on validation and test set using λ (lambda) between 0-1 with intervals of 0.1.*

A weight of 0.2 was found to be optimal on the validation set. Using this weight, the performance of the pretrained ECAPA-TDNN was improved by 3% on the unseen test set.

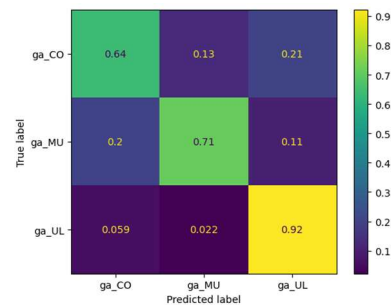

Figure 4: *Confusion matrix of fused acoustic and text-based DID models evaluated using the test set.*

A confusion matrix of the fused systems results is shown in Figure 4. Interestingly, fusing the acoustic and text models helps most with disambiguating between the two southern dialects of Co and Mu with an increase in accuracy of 6% for both, while performance for Ul drops by about 2%.

## 6. Discussion and Conclusion

The experiments yield further insights into the degree of acoustic similarity of the dialects and results echo findings from the dialect bias experiment [4], that on an acoustic level, the Ulster dialect stands apart from Connacht and Munster. Using an acoustic classification model, the Ulster dialect was identified with ~94% accuracy. Performance for Connacht and Munster with was markedly poorer, although it was helped with model fusion. This suggests that to disambiguate these two dialects, more features beyond acoustic and language models may be necessary, such as linguistically salient dialect markers.

When initialised randomly, the ECAPA-TDNN model outperformed XLS-R by a margin of 9%, and by 18% after finetuning. A future direction will explore the usefulness of using self-supervised models as a feature extractor in ECAPA-TDNN training, as well as the incorporation of DID into the ASR process as mentioned previously.

## 7. Acknowledgements

The ABAIR initiative is supported by the Department of Tourism, Culture, Art, Gaeltacht, Sport & Media, with funding from the National Lottery, as part of the 20 Year Strategy for Irish.